\documentclass[letterpaper, 10pt, conference]{ieeeconf} 

\IEEEoverridecommandlockouts
\usepackage[nolist]{acronym}
\usepackage[ruled]{algorithm2e}
\usepackage{amsmath}
\usepackage{bm}
\usepackage{booktabs}
\usepackage{cite}
\usepackage{flushend}
\usepackage{graphicx}
\usepackage[utf8x]{inputenc}
\usepackage{microtype}
\usepackage{newtxmath}
\usepackage{pgfplots}
\usepackage[binary-units=true]{siunitx}
\usepackage{tabularx}
\usepackage{tikz}
\usetikzlibrary{bayesnet}
\usepackage{todonotes}
\makeatletter
\let\NAT@parse\undefined
\makeatother
\usepackage{hyperref}

\newcommand{\aref}[1]{Alg.~\ref{#1}}
\newcommand{\eref}[1]{eq.~\eqref{#1}}
\newcommand{\fref}[1]{Fig.~\ref{#1}}

\newcommand{\tref}[1]{Table~\ref{#1}}
\renewcommand{\vec}[1]{\boldsymbol{\mathbf{#1}}}
\pgfplotsset{compat=newest} 
\pgfplotsset{plot coordinates/math parser=false}
\title{\LARGE \bf Depth Camera Based Particle Filter for Robotic Osteotomy Navigation}

\author{Tim Übelhör$^{1}$, Jonas Gesenhues$^{1}$, Nassim Ayoub$^{2}$, Ali Modabber$^{2}$ and Dirk Abel$^{1}$
  \thanks{*Funded by the Excellence Initiative of the German federal and state governments.}
  \thanks{$^{1}$Institute of Automatic Control, RWTH Aachen, 52074, Germany \newline {\tt\small \{T.Uebelhoer, J.Gesenhues, D.Abel\} @irt.rwth-aachen.de}}
  \thanks{$^{2}$Department of oral and maxillofacial surgery, Uniklinik RWTH Aachen, 52074, Germany {\tt\small \{nayoub, amodabber\}@ukaachen.de}}
}

\begin{document}
\maketitle
\thispagestyle{empty}
\pagestyle{empty}

\begin{acronym}
  \acro{ar}[AR]{augmented reality}
  \acro{cnn}[CNN]{convolutional neural network}
  \acro{drf}[DRF]{dynamic reference frame}
  \acro{pf}[PF]{particle filter}
\end{acronym}

\begin{abstract}
  % Keywords: State Estimation, Robot Vision, Particle Filters, Mobile Robots,Medical Applications
  Active surgical robots lack acceptance in clinical practice, because they do not offer the flexibility and usability required for a versatile usage:
  the systems require a large installation space or a complicated registration step, where the preoperative plan is aligned to the patient and transformed to the base frame of the robot.
  In this paper, a navigation system for robotic osteotomies is designed, which uses the raw depth images from a camera mounted on the flange of a lightweight robot arm.
  Consequently, the system does not require any rigid attachment of the robot or fiducials to the bone and the time-consuming registration step is eliminated.
  Instead, only a coarse initialization is required which improves the usability in surgery.
  The full six dimensional pose of the iliac crest bone is estimated with a particle filter at a maximum rate of \si{90~\hertz}.
  The presented method is robust against changing lighting conditions, blood or tissue on the bone surface and partial occlusions caused by the surgeons.
  Proof of the usability in a clinical environment is successfully provided in a corpse study, where surgeons used an augmented reality osteotomy template, which was aligned to bone via the particle filters pose estimates for the resection of transplants from the iliac crest.
\end{abstract}

\section{Introduction}

  To reconstruct defect jaw bones, transplants from the iliac crest can be resected.
  State of the art is to use 3D printed osteotomy templates as guides to cut out matching pieces from the pelvis.
  These templates have the disadvantages of being fragile, not being customizable during the intervention and inflicting additional damage to the pelvis through the rigid attachment with screws \cite{tack3DprintingTechniquesMedical2016}.
  Robotic surgical systems have the potential to execute surgeries with a high accuracy and reduce the operative time after the setup has been completed \cite{taylorMedicalRoboticsComputerIntegrated2008, riekHealthcareRobotics2017}.
  One possibility to eliminate the additional damage and to offer a higher flexibility is to guide a robot directly via the image data from a camera which is mounted on the flange of the robot.

  To enable the autonomous transplant dissection, the robotic setup has to be able to track the pose of the iliac crest reliably to calculate the appropriate cutting trajectories.
  In this paper, a depth camera based particle filter for the navigation in robotic osteotomies via augmented reality projection is developed, to confirm that a depth camera based navigation system is generally capable of guiding a robot during an osteotomy.

  Currently, the practical applications of robots in surgery are mostly limited to tele-operated soft tissue surgeries, the best known example being the \textit{da Vinci} robot (Intuitive Surgical, Sunnyvale, CA, USA). In this case the control loop of the robot is closed by the visual feedback to the surgeon and his reaction to movements of the tissue \cite{moustrisEvolutionAutonomousSemiautonomous2011, satavaSurgicalRoboticsEarly2002}.
  Autonomously acting surgical robotic platforms for osteotomies are still topic of research and mostly serve as surgical assistants or are restricted to specific applications like craniotomies, stereotaxies or arthroplasties \cite{wolfMedicalAutomationRobotics2009, mezgerNavigationSurgery2013}.
  Only a few systems were launched to the market: for example the \textit{NeuroMate} (Renishaw plc, New Mills, UK) aligns electrodes precisely for stereotactic surgeries and the more recent \textit{CARLO} (AOT AG, Basel CH) uses a cold ablation laser to perform craniotomies autonomously.
  Both systems are used in conjunction with preoperative CAD/CAM systems to perform the preoperative planning and subsequently require the registration of the patient to the robots base frame \cite{cardinaleNewToolTouchfree2017, mezgerNavigationSurgery2013,taylorMedicalRoboticsComputerintegrated2003}.
  During the surgical procedure the patient and robot either have to be fixed statically or their relative poses have to be tracked via \acp{drf} attached to the patient and the robot \cite{kongIntegratedSystemPlanning2016, mezgerNavigationSurgery2013, wolfMedicalAutomationRobotics2009, zhuProspectsRobotAssistedMandibular2016}.
  However, these systems introduce an additional reduction of the available operation room because an additional line of sight between the \acp{drf} and the tracking camera has to be established \cite{liApplicationAccuracyNeuroMate2002, cardinaleNewToolTouchfree2017, schneiderDirectCalibrationLaser2015}. 
  
  A setup consisting of a stereo camera and a mobile projector, both mounted directly on a robot arm, does not require this second line of sight or the attachment of fiducial markers, so the required registration step could be omitted by navigating directly on the basis of the image data.
  This approach introduces several challenges:
  unreliable lighting conditions and textures cause classical object pose estimation algorithms to fail as they require the extraction and matching of keypoints or templates \cite{loweObjectRecognitionLocal1999b, hinterstoisserModelBasedTraining2013}.
  Additionally, the iliac crest is located in a cluttered environment with partial occlusions caused by the surrounding tissue and the operating surgeons.
  More recent \ac{cnn} based methods promise to obtain better results in such challenging environments but require a large amount of annotated data which is not available prior to the operation \cite{wangDenseFusion6DObject2019, brachmannLearning6DObject2014, xiangPoseCNNConvolutionalNeural2017}.
  Moreover, a high update rate and temporal consistency are required to navigate a robot.

  In this paper, these problems are solved by employing a \ac{pf} approach to track the pose of the iliac crest over time.
  The presented algorithm only relies on the depth data of a RGBD-camera, the robots joint encoder measurements and a 3D model of the iliac crest.
  In summary, the main contributions of this work are:
  \begin{itemize}
    \item {A \ac{pf} operating on the raw depth images of a RGBD camera to track the pose of the iliac crest is presented.}
    \item {Practical issues are solved: Gimbal locks are prevented by using a quaternion representation and numerical issues are solved by formulating the \ac{pf} in the logarithmic domain.}
    \item{The system was validated under clinical conditions in a corpse study during the resection of transplants from the iliac crest.}
  \end{itemize}

\section{Related Work}

  Computer vision, especially pose estimation and tracking is a major research field in robotics as it enables robots to interact with a dynamic environment.
  Classical computer vision relies on hand crafted features like SIFT \cite{loweObjectRecognitionLocal1999b} for color or CVFH \cite{aldomaCADmodelRecognition6DOF2011} for depth data.
  These features can be used to match keypoints of a CAD model to the captured image and estimate the pose of the target object by solving the perspective-n-point problem.
  More recent approaches learn the feature representations, for example Brachmann et al.\ train random forests to predict an intermediate representation called \textit{object coordinates} densely for each pixel.
  After the prediction, an energy function, which can be calculated by a \ac{cnn} \cite{krullLearningAnalysisbySynthesis6D2015}, is minimized to regress an objects pose \cite{brachmannLearning6DObject2014}.
  The \textit{PoseCNN} presented by Xiang et al.\ learns to regress poses from RGB images without an intermediate representation \cite{xiangPoseCNNConvolutionalNeural2017}.
  Pose estimates are commonly refined to achieve higher accuracies, for example by using the iterative closest point algorithm or iteratively applying a dedicated refinement \acp{cnn} \cite{liDeepIMDeepIterative2018, wangDenseFusion6DObject2019}.
  Recently it has been shown that \acp{cnn} tend to rely more on textures than shapes, which makes their usage in surgical environments more difficult \cite{geirhosIMAGENETTRAINEDCNNSARE2019}.

  Compared to the pose estimation of an object from a single image, pose tracking algorithms can exploit the temporal relationship between multiple images to improve the estimates.
  Tan et al.\ propose to use random forests to correct the error between the pose estimate from the last and the newly captured image \cite{tanMultiforestTrackerChameleon2014}.
  A \ac{cnn} is used by Garon et al.\ to calculate the pose difference between a synthetically rendered pose and a captured image \cite{garonDeep6DOFTracking2017}.
  As opposed to the learning based approaches, \acp{pf} are a model based approach to probabilistically track the pose of an object with a depth camera and a known CAD model \cite{changhyunchoiRGBDObjectTracking2013, wuthrichProbabilisticObjectTracking2013}.
  
  In clinical applications, the application of recent developments in sensor technology and computer vision is the subject of research.
  For instance, Zhu et al.\ execute the registration by inserting titanium screws as fiducial markers and match them to the corresponding points in the CT data \cite{zhuProspectsRobotAssistedMandibular2016}.
  Attaching a combination of \ac{ar} markers and inertial measurement units rigidly to the bone, can enable robust tracking under occlusions while preventing drifting issues \cite{pflugiCosteffectiveSurgicalNavigation2016, linMandibularAngleSplit2016, pflugiAugmentedMarkerTracking}.
  A non invasive alternative is a point-wise registration of anatomical landmarks using a hand-held pointer probe or custom registration devices \cite{pflugiCosteffectiveSurgicalNavigation2016, kongIntegratedSystemPlanning2016}.
  
  Recent research has focused on improving the usability of surgical navigation systems and reducing the invasiveness \cite{mezgerNavigationSurgery2013,moustrisEvolutionAutonomousSemiautonomous2011}.
  For example Zhang et al.\ use a RGBD camera in conjunction with the \textit{Kinect SDK} to track a face without any fiducials or the selection of anatomical landmarks \cite{zhangKinectbasedAutomaticSpatial2014}.
  For this purpose, they employ the coherent point drift and iterative closest point algorithms to precisely match the surface of preoperative images to the captured images.

  Ma et al.\ propose to use the \ac{pf} implementation from the \textit{point cloud library} to track a patients movements and adjust the renderings of an \ac{ar} overlay device.
  The system requires an exact initialization, runs at \si{10~\hertz} and tracks a template captured from a single view, so the tracking only works for similar views \cite{maAutomaticFastRegistrationSurgical2016}.

\section{Method}
\label{sec:method}

  Due to the aforementioned advantages, in this paper a \ac{pf} is designed to operate on raw depth images to estimate the pose of an iliac crest with a given CAD model.
  The \ac{pf} is a nonparametric Bayesian filter which recursively approximates the probability distribution of the current state, denoted as belief, by a set of particles.
  For every time step a new belief of the bone pose is \textit{predicted} via a transition model and \textit{updated} by incorporating the cameras measurements via an observation model.
  The nonparametric representation is particularly necessary to model the non-Gaussian observations of the depth camera and the resulting multi-modal belief.
  In the next sections the filter equations, the transition model, observation model and a logarithmic formulation of a sequential importance resampling \ac{pf} are derived to probabilistically track the pose of an iliac crest.

  \subsection{Filter Equations}
 
  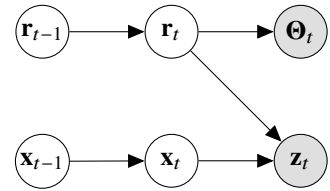
\begin{figure}[t]
    \centering
    \begin{tikzpicture}
  % robot nodes, p = past
  \node[latent](rp){\(\vec{r}_{t-1}\)};
  \node[latent, right=of rp](r){\(\vec{r}_t\)};
  \node[obs, right=of r](t){\(\vec{\Theta}_t\)};
  % object nodes
  \node[latent, below=of rp](xp){\(\vec{x}_{t-1}\)};
  \node[latent, below=of r](x){\(\vec{x}_t\)};
  % camera node
  \node[obs, below=of t](z){\(\vec{z}_t\)};
  % Connections
  \edge{rp}{r}
  \edge{xp}{x};
  \edge{r}{t}
  \edge{r,x}{z};
\end{tikzpicture}
    \caption[system description]{The Bayesian network describing the latent states of the robot \(\vec{r}_t\) and the bone \(\vec{x}_t\) at the time \(t\).
    The robot outputs the measurements \(\vec{\Theta}_t\) from the joint encoders.
    The camera which is mounted on the robots flange observes the depth image \(\vec{z}_t\). }
    \label{fig:bayes_net}
  \end{figure}

  The goal of the filtering problem is to estimate the latent state of the bone \(\vec{x}_t\) at the time \(t\) from a sequence of observations from the robots joint encoders \(\vec{\Theta}_{1:t}\) and the depth camera \(\vec{z}_{1:t}\,\).
  Additionally the cameras observations depend on the latent state of the robot \(\vec{r}_t\), because it is mounted on the flange of the robot.
  The full posterior of the system, i.e. the probability distribution over the states given a sequence of observations, can be separated into an estimation of the bones and the robots posterior.
  With the assumption of \(\vec{x}_t\) being conditionally independent of \(\vec{\Theta}_t\) given \(\vec{r}_t\) (see \fref{fig:bayes_net}) follows:
  \begin{equation}
    p(\vec{r}_t,\vec{x}_t|\vec{\Theta}_{1:t},\vec{z}_{1:t})
    = p(\vec{r}_t|\vec{\Theta}_{1:t},\vec{z}_{1:t}) p(\vec{x}_t|\vec{r}_{t},\vec{z}_{1:t})
    \label{eq:full_posterior}
  \end{equation}
  For the following equations the posteriors are abbreviated as the updated belief:
  \begin{align*}
    bel(\vec{r}_t)&=p(\vec{r}_t|\vec{\Theta}_{1:t},\vec{z}_{1:t}) \\
    bel(\vec{x}_t)&=p(\vec{x}_t|\vec{r}_{t},\vec{z}_{1:t})
  \end{align*}
  Applying Bayes law and the Markov assumption to \eref{eq:full_posterior} leads to the filter equations with the normalization factors \(\eta_r\) and \(\eta_x\):
  \begin{align}
    bel&(\vec{r}_t) = \label{eq:bel_r}\\
    &~\eta_r p(\vec{\Theta}_t|\vec{r}_t) p(\vec{z}_t|\vec{r}_t)
    \int p(\vec{r}_t|\vec{r}_{t-1}) bel(\vec{r}_{t-1} ) ~d\vec{r}_{t-1} \nonumber\\
    bel&(\vec{x}_t) = \label{eq:bel_x}\\
    &~\eta_x \frac{p(\vec{z}_t|\vec{r}_t,\vec{x}_t)} {p(\vec{z}_t|\vec{r}_t)}
    \int p(\vec{x}_t|\vec{x}_{t-1}) bel(\vec{x}_{t-1}) ~d\vec{x}_{t-1} \nonumber
  \end{align}
  The equations above resemble prediction and the update step of a Bayesian filter: the integral terms predict the new belief via the transition model from the old belief and the terms in front of the integrals update the predicted beliefs via the observation model.
  The observation of the camera \(\vec{z}_t\) is the effect of the two causes: \(\vec{r}_t\) and \(\vec{x}_t\).
  Consequently, the likelihood \(p(\vec{z}_t|\vec{r}_t)\)  of measuring a particular depth image given the robots pose appears in both \eref{eq:bel_r} and \eqref{eq:bel_x}.
  Intuitively \(p(\vec{z}_t|\vec{r}_t)\) weights \(bel(\vec{x}_t)\) and \(bel(\vec{r}_t)\):
  A high confidence that an observation is caused by a certain robot pose \(\vec{r}_t\) reduces the probability of the observation being caused by a certain bone pose \(\vec{x}_t\).

  To simplify the problem, the robots joint encoders are assumed to be highly accurate so the robots true state approximately matches the measured state:
  \begin{equation}
    \vec{r}_t \approx \vec{\Theta}_t
  \end{equation}
  Additionally, the position of robots base is assumed to be quasi-static during the operation, which implies \(p(\vec{z}_t|\vec{r}_t) \approx 1\) so the belief of the bones state from \eref{eq:bel_x} becomes:
  \begin{equation}
    bel(\vec{x}_t)
    = \eta_x p(\vec{z}_t|\vec{\Theta}_t,\vec{x}_t)
    \int p(\vec{x}_t|\vec{x}_{t-1}) p(\vec{x}_{t-1}) d\vec{x}_{t-1}
    \label{eq:bel_x_simple}
  \end{equation}
  While \eref{eq:bel_x_simple} does not explicitly account for movements of the robots base or offsets in the robots joint encoders, these effects are corrected implicitly as the estimated bone pose \(\vec{x}_t\) is corrected by the camera measurements \(\vec{z}_t\).
  
  \subsection{Transition Model}
  \label{sec:transition_model}

  The transition model predicts the movement of the bone, which, for example, is caused by surgeons repositioning the patient.
  Therefore, the bones dynamics are described by a decaying velocity model which is perturbed by white noise accelerations \(\vec{w}_t\).
  If no force is continuously applied to the bone, it looses energy due to friction over time.
  The state \(\vec{x}_t\) of the bone is described by the vector \(( \vec{p}_t, \vec{v}_t, \vec{q}_t, \vec{\omega}_t )\):
  the three dimensional cartesian position and velocity as well as the orientation represented as quaternion and the three dimensional angular rate.
  The cartesian position and velocity are described by a time-discrete linear state space model:
  \begin{equation}
    \begin{pmatrix}
      \vec{p}_t \\
      \vec{v}_t \\
    \end{pmatrix} =
    \begin{pmatrix}
      \vec{I} & \Delta T\, \vec{I} \\
      \vec{0} & \lambda_v\, \vec{I}
    \end{pmatrix}
    \begin{pmatrix}
      \vec{p}_{t-1} \\
      \vec{v}_{t-1}
    \end{pmatrix} + 
    \begin{pmatrix}
      \frac{\Delta T^2}{2}\, \vec{I} \\
      \Delta T\, \vec{I}
    \end{pmatrix} \vec{w}_{t,p}
    \label{eq:trans_p}
  \end{equation}
  % \text{with} \lambda_v \in [0,1], w_{k-1} \sim \mathcal{N}(0, q)
  In the equation above \(\Delta T\) is the sample time of the algorithm, \(\vec{I}\) the identity matrix and \(\lambda_v \in [0,1]\) the velocity decay factor.
  The perturbations \(\vec{w}_{t,p}\) are normally distributed, for example with a standard deviation of \si{5~\meter\per\square\second}.
  If the scene is expected to be quasi-static and the tracked object to be occluded frequently, \(\lambda_v=0\) is a reasonable choice because errors in the velocity estimates are not integrated into the state and the estimated pose drifts more slowly.
  However, a lambda of zero leads to only a small region around the current pose being covered by the predicted particles because the current velocity is ignored.  
  Therefore a non-zero lambda, for example \(\lambda_v = 0.95\), is required to enable a stable tracking of fast motions.
  In this case the integration causes the uncertainty to grow quickly with large sample times \(\Delta T\).
  To improve the tracking stability, a small sample time and therefore a high tracking rate is beneficial.

  Quaternions are used to represent the orientation \(\vec{q}_t\) of the bone, because they do not suffer from gimbal locks and are computationally more efficient than Euler angles or rotation matrices.
  However, angular rates \(\vec{\omega}_t\) can not be directly represented as quaternions.
  To derive a transition model which contains both, quaternions and angular rates, consider the time-derivative of a quaternion:
  \begin{equation}
    \dot{\vec{q}}(t) = \frac{1}{2}\vec{q}(t) \otimes \vec{\omega}(t)
    \label{eq:d_quat}
  \end{equation}
  The symbol \(\otimes\) in the equation above denotes the quaternion product.
  Approximating \(\vec{q}_t\) with a Taylor series leads to the discrete backward integration rule \cite{solaQuaternionKinematicsErrorstate2017}:
  \begin{equation}
    \vec{q}_{t} = \vec{q}_{t-1} \otimes e^{\frac{\Delta T}{2} \vec{\omega}_{t}}
    \label{eq:quat_taylor}
  \end{equation}
  The quaternion representation of the incremental rotation \( \Delta T \vec{\omega}_{t}\) can be calculated with the following equation:
  \begin{equation}
    e^{\frac{\Delta T}{2} \vec{\omega}_{t}} =
    \begin{pmatrix}
      \cos\left\lVert \frac{\Delta T}{2} \vec{\omega}_{t} \right\rVert \\
      \frac{\vec{\omega}_{t}}{\left\lVert \vec{\omega}_{t} \right\rVert}
      \sin\left\lVert \frac{\Delta T}{2} \vec{\omega}_{t} \right\rVert
    \end{pmatrix}
    \label{eq:quat_incr}
  \end{equation}
  Since the incremental rotation \(\Delta T \vec{\omega}_{t}\) is a linear approximation, a linear prediction model similar to \eref{eq:trans_p} can be used \cite{QuaternionKalmanFilter}:
  \begin{equation}
    \begin{pmatrix}
      \Delta T \vec{\omega}_{t} \\
      \vec{\omega}_t \\
    \end{pmatrix} =
    \begin{pmatrix}
      \Delta T\,\vec{I} \\
      \lambda_v\,\vec{I} \\
    \end{pmatrix}
    \vec{\omega}_{t-1} + 
    \begin{pmatrix}
      \frac{\Delta T^2}{2}\,\vec{I} \\
      \Delta T\,\vec{I}
    \end{pmatrix} \vec{w}_{t,\omega}
    \label{eq:trans_q}
  \end{equation}
  The parameters \(\Delta T\) and \(\lambda_v \) are the same as in \eref{eq:trans_p}.
  The perturbations \(\vec{w}_{t,\omega}\) are normally distributed, for example with a standard deviation of \si{50~\radian\per\square\second}.
  The result of combining \eref{eq:trans_q} with \eqref{eq:quat_incr} and \eqref{eq:quat_taylor} is a non-linear transition model for the orientation, which is natively supported by the \ac{pf}.

  \subsection{Observation Model}
  \label{sec:observation_model}

  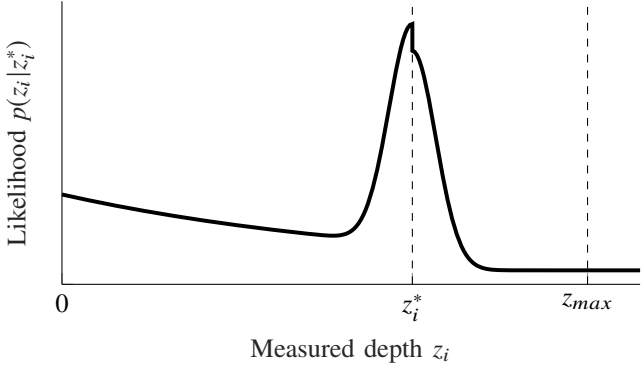
\begin{figure}
    \centering
    % This file was created by matlab2tikz.
%
%The latest updates can be retrieved from
%  http://www.mathworks.com/matlabcentral/fileexchange/22022-matlab2tikz-matlab2tikz
%where you can also make suggestions and rate matlab2tikz.
%
\begin{tikzpicture}

\begin{axis}[%
width=0.89\linewidth,
height=0.436\linewidth,
at={(0\linewidth,0\linewidth)},
scale only axis,
xmin=0,
xmax=2.475,
xtick={0,1.5,2.25},
xticklabels={{0},{$z_i^*$},{$z_{max}$}},
xlabel style={font=\color{white!15!black}},
xlabel={Measured depth $z_i$},
ymin=0,
ymax=1.8,
ytick={0},
yticklabels={{}},
ylabel style={font=\color{white!15!black}},
ylabel={Likelihood $p(z_i|z_i^*)$},
axis background/.style={fill=white},
axis x line*=bottom,
axis y line*=left
]
\addplot [color=black, line width=1.5pt, forget plot]
  table[row sep=crcr]{%
0	0.571397771622817\\
0.11	0.535975859386208\\
0.23	0.500292797895809\\
0.35	0.467457685375857\\
0.48	0.434836961194174\\
0.61	0.405027117507693\\
0.75	0.375790387089341\\
0.9	0.347459018130609\\
1.05	0.321981261413582\\
1.12	0.311910060733442\\
1.15	0.309373955210443\\
1.17	0.309355124683532\\
1.19	0.311807779180773\\
1.21	0.318296199138225\\
1.22	0.323725694174092\\
1.23	0.331064115242828\\
1.24	0.340710348520906\\
1.25	0.353107935525896\\
1.26	0.36873841441483\\
1.27	0.388110544724661\\
1.28	0.411744954636437\\
1.29	0.440153979050632\\
1.3	0.473816769816949\\
1.31	0.513150147604656\\
1.33	0.609987369683325\\
1.35	0.731484871962995\\
1.37	0.875357425054285\\
1.4	1.11862310146795\\
1.43	1.36085160803665\\
1.45	1.4977261209083\\
1.46	1.55322304366504\\
1.47	1.59792340872034\\
1.48	1.63051235395563\\
1.49	1.65000202226707\\
1.5	1.65577952177586\\
1.5	1.4851868702939\\
1.51	1.47822280505843\\
1.52	1.45753831780298\\
1.53	1.42374624300072\\
1.54	1.37783437995052\\
1.55	1.32111753256394\\
1.57	1.18177765567255\\
1.6	0.93578642470589\\
1.63	0.688678961056215\\
1.65	0.542200473719511\\
1.67	0.418060659707994\\
1.69	0.318544240600257\\
1.71	0.242831474820384\\
1.72	0.213049963850699\\
1.73	0.188033520984493\\
1.74	0.167269744920839\\
1.75	0.150237940616379\\
1.76	0.136429281206789\\
1.77	0.125362160739368\\
1.78	0.116592969429319\\
1.8	0.104400358330672\\
1.82	0.0972331975940159\\
1.84	0.0932016559785445\\
1.87	0.0903756198358163\\
1.92	0.0890951896260366\\
2.14	0.0888888906697378\\
2.475	0.0888888888897408\\
};
\addplot [color=black, dashed, forget plot]
  table[row sep=crcr]{%
1.5	0\\
1.5	1.8\\
};
\addplot [color=black, dashed, forget plot]
  table[row sep=crcr]{%
2.25	0\\
2.25	1.8\\
};
\end{axis}
\end{tikzpicture}%
    \caption{Qualitative observation model for a single depth pixel. The model is a superposition of an exponential, a normal and a uniform distribution.}
    \label{fig:beam_model}
  \end{figure}

  A beam model similar to  \cite{wuthrichProbabilisticObjectTracking2013} is used, which interprets each pixel \(i\) as a beam hitting an object at the distance \(z_i\).
  The likelihood of measuring a depth of \(z_i\) given an expected depth of \(z_i^*\) is visualized in \fref{fig:beam_model} and modeled by the following equation:
  \begin{equation}
    p\left(z_i|z_i^*\right) = w_u p_u + w_e p_e\left(z_i|z_i^*\right) + w_n p_n\left(z_i|z_i^*\right)
    \label{eq:superposition}
  \end{equation}
  
  The model is a superposition of three probability density functions: a uniform distribution \(p_u\), an exponential distribution \(p_e\) and a normal distribution \(p_n\).
  As stated by Wüthrich et al.\ \cite{wuthrichNewPerspectiveExtension2015}, the algorithm is insensitive to the selection of the weight parameters \(w_u, w_e \text{ and } w_n\).
  As long as all of the weights are non-zero, the tracking is stable.
  First, the uniform distribution handles random measurements like invalid depth values or outliers.
  It is parametrized by the maximum measuring distance of the camera:
  \begin{equation}
    p_u = \frac{1}{z_{max}}
  \end{equation}
  Note that \(p_u\) is theoretically constrained to the maximum measuring distance \(z_{max}\) but as outliers might occur out of this range the same probability is assigned to values exceeding the maximum (see \fref{fig:beam_model}).
  Second, the probability of occlusions is formulated implicitly like proposed by Thrun et al. to save computation resources and enable higher tracking rates \cite{thrunProbabilisticRobotics2005}.
  The probability of observing a random object which occludes the bone decreases exponentially with the depth \(z_i\):
  \begin{equation}
    p_e \left( z_i|z_i^* \right) =
    \begin{cases}
      \frac{\lambda_e e^{-\lambda_e z_i}}{1-e^{-\lambda_e z_i^*}} & 0 < z_i < z_i^* \\
      0 & \text{otherwise}
    \end{cases}
  \end{equation}
  The exponential decay constant \(\lambda_e\) is chosen so that the probability of observing an occluding object is halved every meter.
  Since an occlusion can only occur in front of the bone, the exponential distribution is limited and normalized to the interval \([0;z_i^*]\).
  Third, a normal distribution with the expected depth \(z_i^*\) as mean is used to model the probability of observing the bone:
  \begin{equation}
    p_n \left( z_i|z_i^* \right) = 
    \begin{cases}
      \eta \mathcal{N} \left(z_i; z_i^*, \left(\sigma(z_i^*)\right)^2\right) & 0 < z_i < z_i^* \\
      0 & \text{otherwise}
    \end{cases}
    \label{eq:normal_dist}
  \end{equation}
  It can be shown that the standard deviation \(\sigma(z_i)\) of depth measurements from stereo cameras grows quadratically with the distance \cite{chatterjeeNoiseStructuredLightStereo2015}.
  Additionally some base noise is added since overestimating the noise is beneficial for robust filtering \cite{thrunProbabilisticRobotics2005}:
  \begin{equation}
    \sigma\left(z_i^*\right) = k \left(z_i^*\right)^2 + \sigma_{base}
  \end{equation}
  The standard deviation \(\sigma\left(z_i^*\right)\) is in the order of a few millimeters and very small compared to the interval \([0; z_i^*]\).
  Therefore, the normalization constant \(\eta\) from \eref{eq:normal_dist} can be approximated as one to save computation power.

  \begin{figure}
    \centering
    \includegraphics[width=\linewidth]{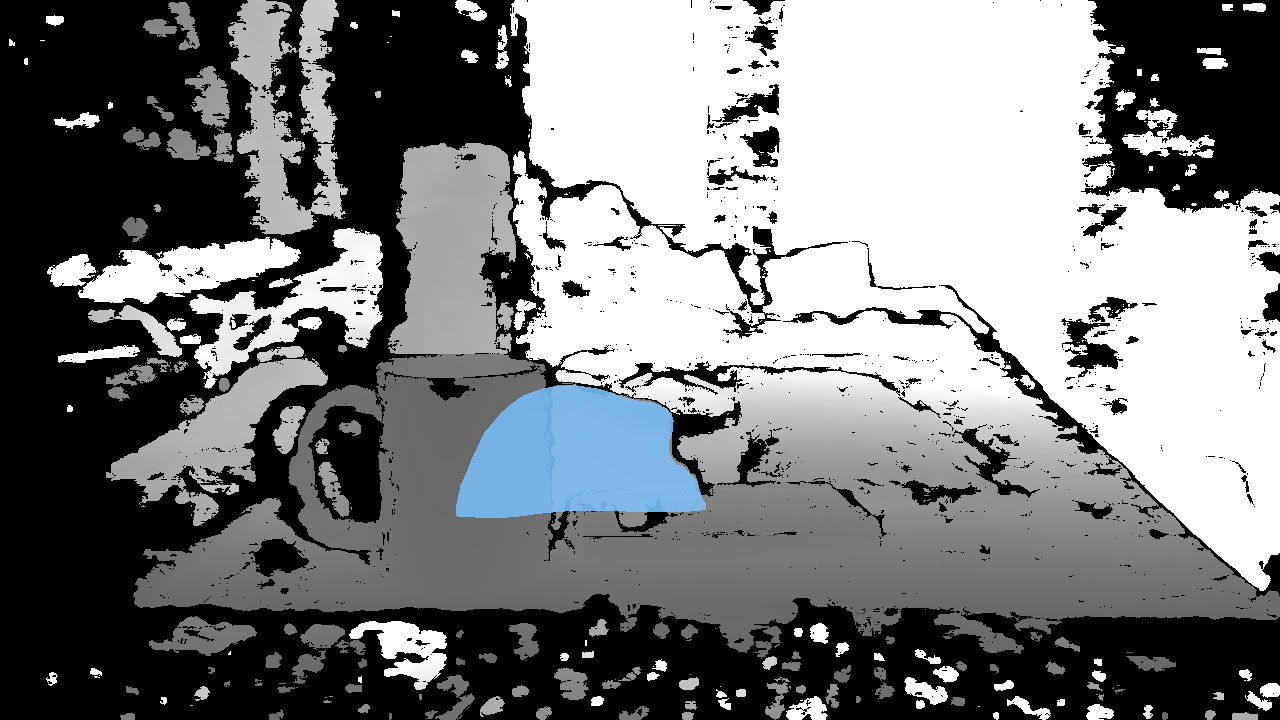}
    \caption{A measured depth image \(\vec{z}_t\) and a rendered depth image \(\vec{z}_t^*\) for the pose hypothesis \(\vec{x}_t\) overlayed in blue. In this picture a 3D printed model of an iliac is placed in a cluttered scene and partially occluded by a calculator and a cup.}
    \label{fig:depth_overlay}
  \end{figure}

  Finally, the likelihood of a pose hypothesis has to be estimated not only for a single pixel but for the whole image.
  For every pixel of the measured image \(\vec{z}_t\) the likelihood is computed via \eref{eq:superposition}, by using the expected depth from a rendered depth image \(\vec{z}_t^*\) for the pose hypothesis \(\vec{x}_t\).
  To render an accurate depth image, a calibration of the cameras extrinsics and intrinsics has to be executed, for example via OpenCV and a ChArUco board \cite{opencv_library}.
  Assuming that each pixel is an independent measurement, the likelihood of the image is the product of all pixel likelihoods:
  \begin{equation}
    p\left(\vec{z}_t|\vec{\Theta}_t, \vec{x}_t\right) = p\left(\vec{z}_t|\vec{z}_t^*\right) = \prod_{i=1}^{N_p}p(z_{i,t}|z_{i,t}^*)
    \label{eq:accum_likely}
  \end{equation}
  Two problems arise from this formulation:
  First, many of the rendered pixels have a depth of zero since the CAD model only covers a small region of the image (compare \fref{fig:depth_overlay}).
  The solution is to assign a likelihood of one to pixels with an expected depth of zero, as any measurement is good for these pixels.
  Second, many pixel likelihoods are much smaller than one and the multiplication of many small values results in a cumulative likelihood of zero because of the limited precision of the computer.
  This numerical issue can be solved by calculating the log-likelihoods instead. Consequently the values are scaled and the product from \eref{eq:accum_likely} becomes a sum:
  \begin{equation}
    \ln \left( p\left(\vec{z}_t|\vec{\Theta}_t, \vec{x}_t\right) \right) = \sum_{i=1}^{N_p}\ln \left(p(z_{i,t}|z_{i,t}^*)\right)
  \end{equation}
  To enable a high tracking rate, the calculation of the log-likelihoods is carried out highly parallelized on a GPU.
  Similar to \cite{wuthrichProbabilisticObjectTracking2013}, all of the pose hypotheses are rendered into one large texture with different viewports via OpenGL, which automatically handles self occlusions in the depth buffer.
  The calculation of the likelihoods is executed in a compute shader: one work group is dispatched for every rendered pose and 128 threads are launched locally in each work group to calculate the pixels log-likelihoods.
  
  \subsection{Logarithmic Particle Filter}

  Numerical issues do not only result from the high dimensional observations but also from the high dimensional state space.
  This leads to many particles having a weight close to zero and causes problems during the weight normalization and resampling step of the \ac{pf}.
  Using logarithmic weights instead leads to a higher numerical stability of the \ac{pf} algorithm.
  Thus, a modified version of the log-\ac{pf}, proposed by Gentner et al.\ \cite{gentnerLogPFParticleFiltering2018}, is used.
  In this formulation the weights are stored logarithmically as \(\bar{w}_{j,t} = \ln(w_{j,t})\) and the weight update and resampling steps are carried out in logarithmic space.

  Given a set of \(N_p\) particles with the states \(\vec{x}_{j,t}\) and logarithmic weights \(\bar{w}_{j,t-1}\), the update from \eref{eq:bel_x_simple} is divided into a weight update and a normalization step.
  The weight update in logarithmic domain turns the product into a sum:
  \begin{align}
    \bar{w}_{j,t} 
    &= \ln \left( w_{j,t-1}~p(\vec{z}_t|\vec{\Theta}_t, \vec{x}_t) \right) \nonumber\\
    &= \bar{w}_{j,t-1} + \ln \left( p(\vec{z}_t|\vec{\Theta}_t, \vec{x}_t) \right)
  \end{align}
  The weight normalization requires calculating the sum of the weights, which is commonly approximated by the log-sum-exp (LSE) function to avoid underflows:
  \begin{align}
    \bar{w}_{j,t}
    &\gets \ln\left( \frac{w_{j,t}}{\sum_{j=1}^{N_w}w_{j,t}} \right) \nonumber\\
    &\approx \bar{w}_{j,t} - LSE \left( \bar{w}_{1,t}~,\dots,~\bar{w}_{N_w,t} \right)
  \end{align}
  Finally, the high dimensionality of the state space causes the particles to degenerate quickly so they have to be resampled frequently.
  The resampling step is carried out with a systematic resampling scheme in the logarithmic domain, see \aref{alg:1}.
  By selecting a random starting point in the interval \([0;N_p^{-1}]\) and systematically iterating over the weights, the sampling variance and computational complexity can be minimized \cite{thrunProbabilisticRobotics2005}.
  \begin{algorithm}[t]
    \caption{Systematic Resampling in Log-Domain.}
    \label{alg:1}
    \KwResult{resampled particle states \(\hat{x}\) and logarithmic weights \(\hat{w}\)}  
    \KwIn{particle states \(x\) and logarithmic weights \(\bar{w}\)}
    \SetKwArray{states}{\(x\)}
    \SetKwArray{weights}{\(\bar{w}\)}
    \SetKwArray{newstates}{\(\hat{x}\)}
    \SetKwArray{newweights}{\(\hat{\hat{w}}\)}
    \(c \gets \weights{\,0\,}\); \tcp{cumulative log-weight}
    \(r \gets \text{random}(0,1/N_p)\); \tcp{starting point}
    \(i \gets 1\); \tcp{current sample}
    \For{\(n \gets 0 \ \KwTo \ N_s\)}
    {
      \tcp{sample from position U}
      \(U \gets \ln \left( r + n/N_p \right)\)\;
      \While{\(U > c\)}
      {
        \(i \gets i + 1\)\;
        \tcp{Jacobian logarithm}
        \(c \gets \max \left( c,\weights{\,i\,}\, \right) + \ln \left( 1 + \exp(-|\,c-\weights{\,i\,}\,|\,) \right) \);
      }
      $\newstates{\,n\,} \gets \states{\,i\,}$\;
      $\newweights{\,n\,} \gets 1/N_p$\;        
    }
  \end{algorithm}

\section{Experimental Setup}
\label{sec:experiments}

  \begin{figure}
    \centering
    \includegraphics[width=\linewidth]{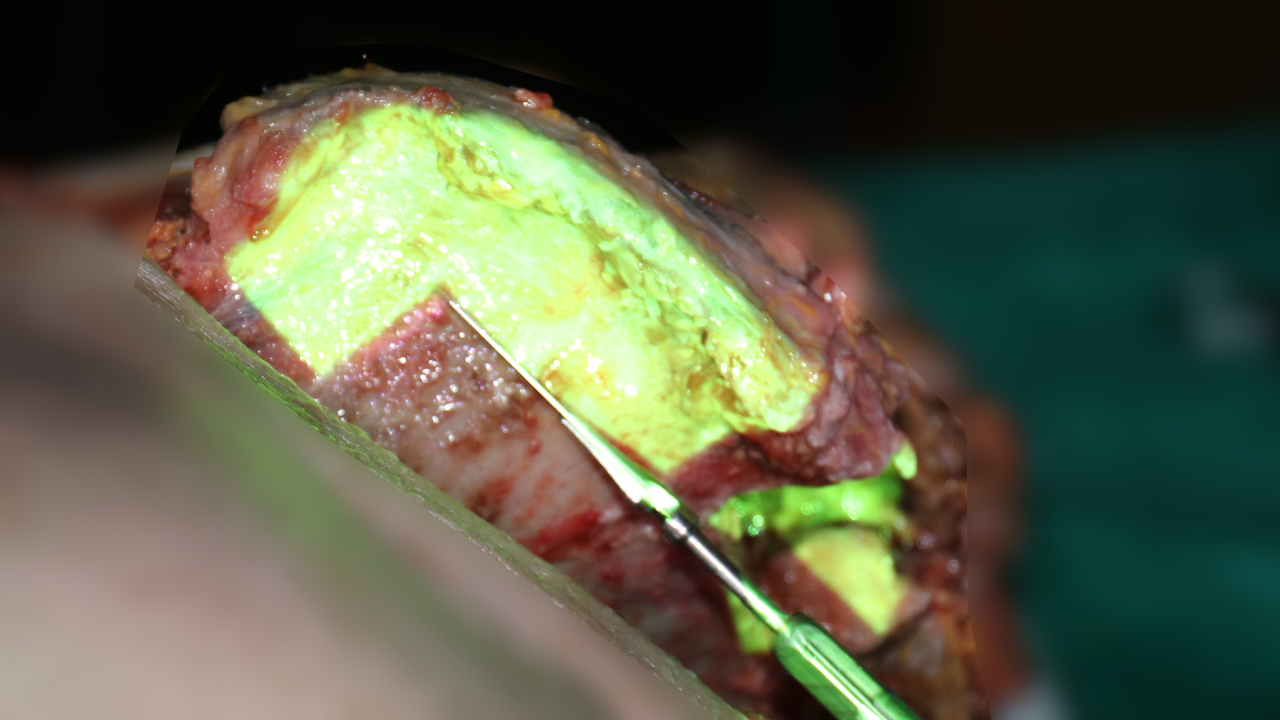}
    \caption{Resection of the hockey (left) and box (right) \ac{ar} templates projected on the iliac crest of a corpse. Note that the box transplant has already been removed and the projection of the hockey remains stable.}
    \label{fig:projection}
  \end{figure}

  Experiments were conducted to validate that the developed \ac{pf} is generally capable of reliably tracking the pose of an iliac crest during an osteotomy.
  As a preliminary step of guiding an actively cutting robot, an \ac{ar} template was projected on the iliac crest by the pose estimates of the presented \ac{pf}.
  This \ac{ar} template was used by surgeons from the department of oral and maxillofacial surgery, Aachen university hospital, to manually resect transplants for the mandibular reconstruction (see \fref{fig:projection}).
  For this purpose, a portable \textit{Optoma ML750ST} projector was mounted above an \textit{Intel Realsense D415} camera on the flange of a \textit{KUKA LRB iiwa R820}.

  The experiments were prepared by performing a CT scan of the pelvis, cleaning the data and transferring it into a planning software.
  In the planning software, a 3D model of the iliac crest was extracted together with a 3D model of the planned transplant which served as model for the \ac{ar} projection.
  An additional model had to be constructed via CAD to serve as conventional physical template and was then manufactured with a 3D printer.
  The \ac{ar} templates were used in one half of the experiments, the other half was executed with conventional 3D printed templates which were screwed to the bones.

  In the first series of experiments, iliac crest models were 3D printed from the CT scans of the pelvis from five real patients.
  For the left and the right iliac crest, a hockey and a box shaped template was planned.
  Consequently 20 different transplants were cut out via the \ac{ar} and the conventional template respectively.
  During the plastic bone experiments, the \ac{pf} was executed on the CPU with 200 particles on downsampled images with a size of \si{80x60~px}.

  In the second series of experiments, the transplants were cut out from five corpses.
  Opposed to the first series, where only one transplant was resected per iliac crest, two transplants were resected from each iliac crest of a corpse.
  For every patient one of the iliac crests was resected via the \ac{ar} templates and the other one via the conventional templates.
  In total, ten transplants were cut out via the \ac{ar} templates and ten transplants were cut out via the conventional templates.
  For both template types, the time required from aligning the template until successfully removing the transplant was measured.
  To ensure equivalent starting conditions, the dissection of the bones was executed before starting the time measurements.
  Procedure specific preparations are included in the time measurements.
  The \ac{ar} template experiments include a manual initialization step using ROS interactive markers which are coarsely aligned to the measured point cloud.
  The conventional templates require pre-drilling of holes and the attachment with screws.
  For the corpse study a GPU implementation was added and the \ac{pf} ran with 1500 particles on depth images of \si{160x120~px} at \si{30 \hertz}.

  The \ac{pf}, the rendering of the \ac{ar} template and data recording were executed on a notebook with an Intel Core i7-7700HQ CPU, \si{16\giga\byte} of RAM and a Nvidia GeForce 940MX GPU.

  % Experiments were conducted to validate the functionality of the system: a quantitative comparison between the \ac{pf} and an ArUco marker baseline as well as a qualitative usability comparison between a \ac{pf} based \ac{ar} templates and conventional templates.
  % \todo[inline]{Angesichts page count weglassen?}
  % In the first series of experiments the \ac{pf} was compared to an ArUco marker \cite{garrido-juradoAutomaticGenerationDetection2014} baseline, to confirm that the pose estimations are accurate enough for an \ac{ar} application. 
  % Four markers have been rigidly attached to a 3D printed iliac and \textit{fiducial\textunderscore slam} ROS-package was used to generate a map of the markers and resolve the redundancy.
  % Compared to the \ac{pf} the marker poses are calculated independently for every image, so they do not suffer from tracking losses in local minima.
  % For the registration of the iliac in the markers frame, the \ac{icp} algorithm was used.

  \section{Results and Discussion}

  \begin{table}[t]
    \centering
    \normalsize
    \begin{tabular}{c c c c c}
      \toprule
        & AR box & AR hockey & Con. box & Con. hockey \\
        & [\si{\second}] & [\si{\second}] & [\si{\second}] & [\si{\second}] \\
      \hline
      min   & 136   & 66    & 105   & 137   \\  
      mean  & 232.2 & 132   & 144.5 & 166.8 \\
      max   & 377   & 208   & 234   & 188   \\
      \bottomrule
    \end{tabular}
    \caption{Time measurements for corps study via \ac{ar} templates and conventional templates. For each template type, box and hockey shaped transplants were cut out.}
    \label{tab:times}
  \end{table}

  % \begin{figure}
  %   \centering
  %   \input{human_times}
  %   \caption{Time measurements for corps study via \ac{ar} templates and conventional templates. For each template type, box and hockey shaped transplants were cut out.}
  %   \label{fig:boxplot}
  % \end{figure}

  During the plastic bone experiments 20 transplants were successfully cut out using the \ac{ar} templates and 20 using the conventional templates.
  The \ac{pf} had proven to be generally capable of tracking the pose of an iliac crest during the resection, with partial occlusions (see \fref{fig:depth_overlay} and \fref{fig:projection}).
  However, in these preliminary experiments one major issues became apparent: the particle filter failed frequently when the bones were reorientated and the \ac{pf} had to be reinitialized manually.
  One possible explanation of the failures is that the predicted particle cloud was too small to track larger pose offsets between two camera images.
  However, increasing the perturbations of the transition model in \eref{eq:trans_p} and \eqref{eq:trans_q} made the tracking unstable in static scenes.
  This behavior could be explained by the fact that the cloud of 200 particles became to sparse to estimate the true pose.
  As a consequence, a GPU implementation was developed to allow larger particle clouds by increasing the number of particles.

  The GPU implementation was used in the corpse study and has proven to be more stable with only a few tracking failures during the resection of the bones.
  Even though the GPU implementation of the \ac{pf} can run at \si{90 \hertz} with 700 particles on images with a size of \si{100x75~px}, a lower frame rate of \si{30 \hertz} had to be used during the experiments as the data recorder caused a bottleneck.
  Nevertheless, ten transplants were successfully lifted from five iliac crests using the \ac{ar} templates and another ten using the conventional templates.
  As illustrated in \fref{fig:projection}, the tracking remained stable even after the first transplant had been removed and partial occlusions caused by the surrounding tissue and the surgeon.
  These observations reinforce the assumption that all three probability distributions from \eref{eq:superposition} are required to enable a stable tracking: the exponential distribution covers occlusions, the normal distribution the expected depths and the uniform distribution outliers like the removed piec.

  The results of the time measurements of the procedures show that transplants can be cut out as quickly as with the conventional templates (see \tref{tab:times}).
  However, some of the \ac{pf} based procedures have taken significantly longer due to the manual initialization of the algorithm.
  Even though only a coarse initialization is required, as the \ac{pf} converges over time, it can be difficult to distinguish the bone from the surrounding tissue in the point cloud.
  A possible solution could be, to initialize the system by coarsely aligning the projected template to the physical bone using the impedance control of the KUKA LBR iiwa.

  In terms of usability, the real benefit of the developed system becomes apparent when the preparation time spent on the CAD construction of the templates is considered.
  The preparation of the \ac{ar} templates is finished after specifying the osteotomy planes, cutting the 3D model with these planes and exporting the transplant as 3D model.
  After this step the conventional templates require up to two hours of construction work.
  Additional time is required for 3D printing and delivering the sterile templates.
  In contrast the \ac{pf} in conjunction with \ac{ar} templates enables rapid adjustments of the planned geometries.

  % TODO
  % On the other hand the parameters of the transition model have to be tuned carefully, depending on the expected dynamics and occlusions:
  % Small \(\lambda_v\) lead to a slower drifting when the bone is fully occluded but the tracking fails under quick motions.
  % Specifically during the bone cutting procedure full occlusions have to be expected but the bone is only repositioned slightly so a small \(\lambda_v\) is a reasonable choice.
  % On the other hand the observation model is robust against almost any parameter changes as long as the sensor noise is overestimated and the weights are non zero.
  % Moreover the depth based observation model is not affected by changing textures or lightning conditions at all.
  % High frame rates of \si{90~\hertz} and particle numbers of \si{700} have proven to be beneficial for tracking stability.
  
\section{Conclusion}

  The depth camera based \acf{pf} has proven to be a suitable solution for the navigation in osteotomies.
  As the depth images offer sufficient geometrical information, no fiducial markers are required which inflict additional damage to the bone and require an additional registration step.
  Moreover, the \ac{pf} seamlessly integrates into the workflow of the surgeons, as the required 3D model is easily exported from the planning tool and no physical templates need to be manufactured.
  The presented work has shown, that the usability of the \ac{pf} in conjunction with an projected \ac{ar} template is equal to the usability of conventional 3D printed templates.
  The \ac{pf} allows a temporal consistent tracking of a bones pose at a rate of \si{90\hertz} which could allow to control an actively cutting robot in future works.
  Topics of future research include the automatic initializations using a pose estimation algorithm \cite{xiangPoseCNNConvolutionalNeural2017, wangDenseFusion6DObject2019} to replace the error-prone manual initialization step and further improve the usability and reduce the operative time.
  Additionally failures of the pose tracking have to be detected automatically to guarantee the safe usage of an actively cutting robot.

  % \begin{itemize}
  %   \item {careful tuning of transition model}
  %   \item {whatever tuning of observation model}
  %   \item {GPU implementation - what? how?}
  %   \item{their framerate 10fps}
  %   \item {handling (self) occlusions? Thin walls -> icp/ full model fails}
  %   \item {167 particles}
  %   \item {need really precise initialization, strictly align, otherwise constant offset}
  %   \item {rely on segmented tracking template from segmentation which is noisy. Additionally only one view is captured so when the camera or patient is moved the point cloud does not match the view.}
  %   \item {overlay device large and static -> limited viewing area}
  %   \item {include system knowledge and robot dynamics}
  %   \item {high dimensionality}
  % \end{itemize}

% \newpage
\bibliographystyle{IEEEtran}
\bibliography{main}

\end{document}